\documentclass[letterpaper]{article} 
\usepackage{aaai24}  
\usepackage{times}  
\usepackage{helvet}  
\usepackage{courier}  
\usepackage[hyphens]{url}  
\usepackage{graphicx} 
\usepackage{natbib}  
\usepackage{caption} 
\usepackage{algorithm}
\usepackage{algorithmic}

\usepackage{xcolor}
\definecolor{mantis}{HTML}{7BC950}
\definecolor{ut_orange}{HTML}{FF8811}
\definecolor{raspberry}{HTML}{D81E5B}
\usepackage{amsmath}
\usepackage{amsfonts}

\usepackage{newfloat}
\usepackage{listings}
\DeclareCaptionStyle{ruled}{labelfont=normalfont,labelsep=colon,strut=off} 
\floatstyle{ruled}
\newfloat{listing}{tb}{lst}{}
\floatname{listing}{Listing}
\title{LAMPAT: Low-Rank Adaption for Multilingual Paraphrasing Using Adversarial Training}
\author{
    Khoi M. Le\textsuperscript{\rm 1, \rm 2}\equalcontrib, 
    Trinh Pham\textsuperscript{\rm 2}\equalcontrib,
    Tho Quan\textsuperscript{\rm 2},
    Anh Tuan Luu\textsuperscript{\rm 3\thanks{Corresponding author.}}
}
\affiliations{
    \textsuperscript{\rm 1}VinAI Research, Vietnam\\
    \textsuperscript{\rm 2}Ho Chi Minh City University of Technology (HCMUT), VNU-HCM, Ho Chi Minh City, Vietnam\\
    \textsuperscript{\rm 3}Nanyang Technological University, Singapore\\
    v.khoilm1@vinai.io, phkhanhtrinh23@gmail.com, qttho@hcmut.edu.vn, anhtuan.luu@ntu.edu.sg
}

\begin{document}

\maketitle

\begin{abstract}
Paraphrases are texts that convey the same meaning while using different words or sentence structures. It can be used as an automatic data augmentation tool for many Natural Language Processing tasks, especially when dealing with low-resource languages, where data shortage is a significant problem. To generate a paraphrase in multilingual settings, previous studies have leveraged the knowledge from the machine translation field, i.e., forming a paraphrase through zero-shot machine translation in the same language. Despite good performance on human evaluation, those methods still require parallel translation datasets, thus making them inapplicable to languages that do not have parallel corpora. To mitigate that problem, we proposed the first unsupervised multilingual paraphrasing model, LAMPAT (\textbf{L}ow-rank \textbf{A}daptation for \textbf{M}ultilingual \textbf{P}araphrasing using \textbf{A}dversarial \textbf{T}raining), by which monolingual dataset is sufficient enough to generate a human-like and diverse sentence. Throughout the experiments, we found out that our method not only works well for English but can generalize on unseen languages as well. Data and code are available at \url{https://github.com/VinAIResearch/LAMPAT}.
\end{abstract}

\section{Introduction}

Paraphrase generation involves the transformation of a given sentence or phrase into its equivalent form while preserving its semantic content. By leveraging the power of NLP techniques, paraphrase generation can enhance several applications, such as machine translation \citep{freitag-etal-2020-human}, information retrieval \citep{NEURIPS2020_d6f1dd03}, question-answering systems \citep{gan-ng-2019-improving}, and text summarization \citep{Cao2017}. However, most existing approaches in paraphrase generation focus on a single language such as English, limiting their effectiveness in multilingual scenarios where accurate and contextually appropriate paraphrases are crucial.

\begin{figure}[h]
    \centering
    \includegraphics[width=0.45\textwidth]{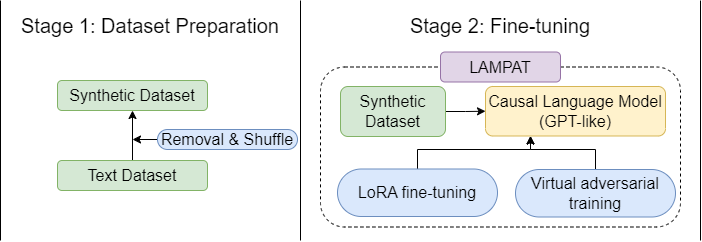}
    \caption{The training process of LAMPAT consists of multiple stages. Firstly, we create a synthetic parallel corpus using unsupervised monolingual data. Next, we utilize LoRA to effectively fine-tune our model. Finally, we obtain the self-supervised model LAMPAT through the utilization of Virtual Adversarial Training.}
    \label{fig:lampat-architecture}
\end{figure}

Most of the current approaches for multilingual paraphrasing are built around the mechanism of machine translation. \citet{thompson-post-2020-paraphrase} make use of the multilingual neural machine translation (MNMT) model from \citet{thompson-post-2020-automatic} to translate between the same source and target languages with a special decoding algorithm to reduce the lexical overlaps, creating a paraphrase of the input text. Meanwhile, round-trip translation \citep{federmann-etal-2019-multilingual}, which creates a pivot language for source language to translate forth and back to create a different wording output, is another approach for paraphrase generation. 

While the machine translation approach is widely recognized for its effectiveness in multilingual paraphrasing, it encounters various limitations. One primary hurdle is the necessity of obtaining reliable parallel corpora of high quality for machine translation, which can be challenging to obtain in real-world scenarios. Another drawback lies in the lack of diversity in the generated output, often resulting in output sequences that closely resemble the input and fail to preserve crucial information present in the original input, as depicted in Table \ref{tab:fail_example_mt}. Many MNMT models utilize heuristic blocking techniques during inference to avoid generating output sequences that are identical to the input. However, this approach limits the model's ability to alter word order or employ diverse syntactic structures, such as inversion or active-to-passive transformations. Consequently, the generation lacks diversity. Furthermore, these blocking algorithms heavily rely on the distribution of the vocabulary. For instance, \citet{thompson-post-2020-paraphrase} reduce the probability of selecting subsequent subwords in $n$-grams to encourage the model to choose different subwords. Nevertheless, these alternative subwords may include antonyms or unrelated words, potentially leading to paraphrases with semantic meanings deviating from the intended direction and producing inappropriate results.

\begin{table}[ht]
    \centering
    {\fontsize{9pt}{9pt}\selectfont
    \begin{tabular}{l|l}
        \textbf{Type} & \textbf{Sentence} \\ \hline
        Input &  I like to eat pasta. \\
        Human reference & \textbf{My favourite food is} pasta. \\
        Lack of diversity & I \textit{like eating} pasta. \\
        Incorrect meaning & I like to eat \underline{paste}. \\
    \end{tabular}}
    \caption{While human can change the structure of the sentence to create paraphrase (in bold), paraphrasing model usually tends to replace words or slightly modify the syntax (in italic). In the worst case, paraphrasing model even changes the meaning of the sentence by using inappropriate word replacement (in underline).}
    \label{tab:fail_example_mt}
\end{table}

\begin{figure*}[h]
    \centering
    \includegraphics[width=0.72\linewidth]{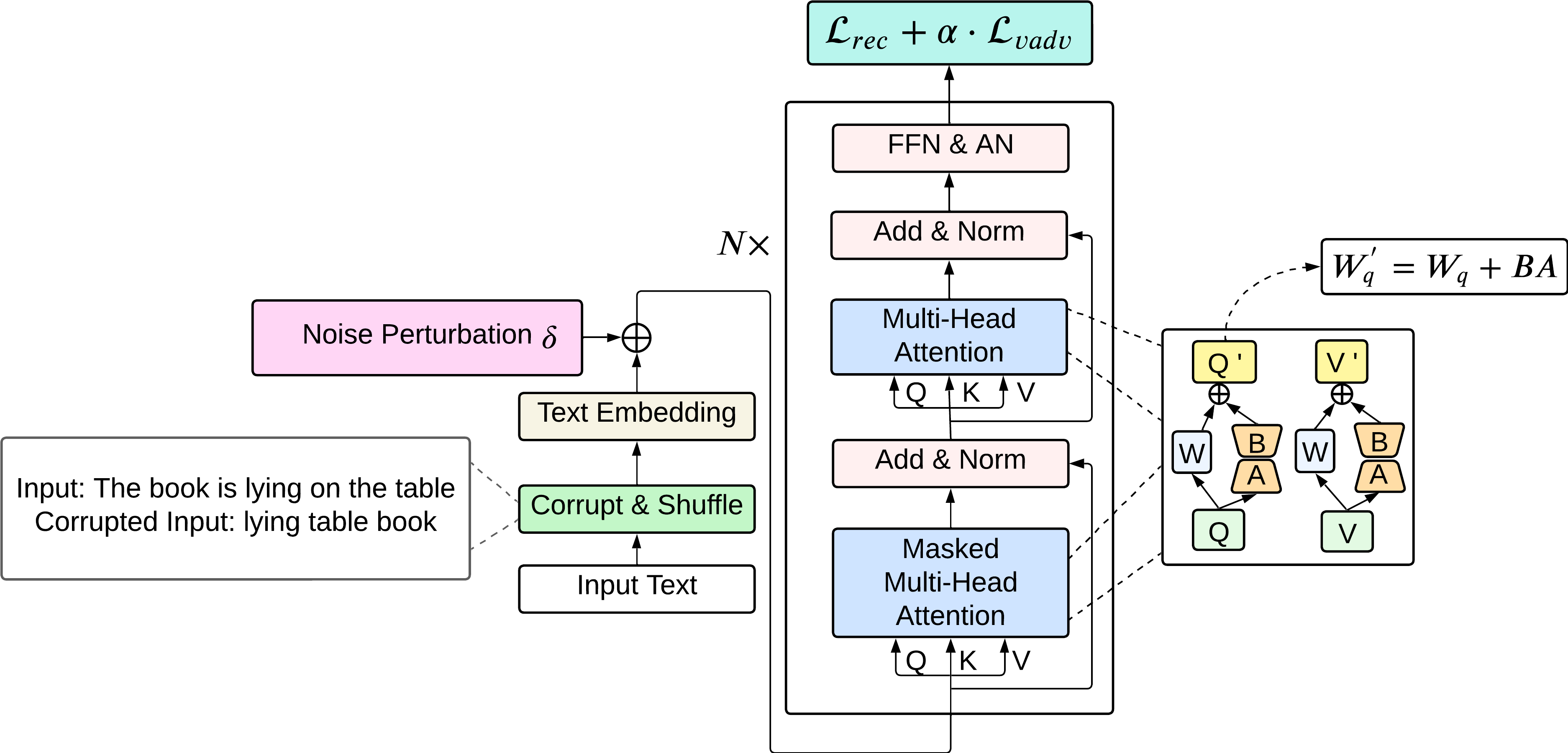}
    \caption{LAMPAT is showcased using actual inputs. Initially, an input text undergoes corruption by removing stopwords and shuffling. Then, a noise perturbation, denoted as $\delta$, is introduced into the text embedding to generate a paraphrase that exhibits lexical diversity. The transformer block is replicated $N$ times, with the Multi-Head Attention component decomposed into low-rank matrices for efficient fine-tuning. Lastly, LAMPAT is trained using virtual adversarial training, incorporating a two-component loss function: the reconstruction loss $\mathcal{L}_{rec}$ and the virtual adversarial regularizer $\mathcal{L}_{vadv}$.}
    \label{fig:detail-lampat-architecture}
\end{figure*}

In this work, we propose LAMPAT (\textbf{L}ow-rank \textbf{A}daptation for \textbf{M}ultilingual \textbf{P}araphrasing using \textbf{A}dversarial \textbf{T}raining) as an approach to mitigate the strict requirements of parallel corpora by learning in an unsupervised manner and alleviate the problem of duplicate generation by using adversarial training objectives. According to Figure \ref{fig:lampat-architecture}, to eliminate the need for parallel corpora, we use the monolingual dataset and apply a series of processes (i) identify stop words (ii) remove stop words (iii) randomly shuffle the words to create the corrupted version of the original input. The training of the model focuses on the objective of reconstructing the original sentence from a corrupted version, aiming to recreate the initial sentence. However, by this learning objective, the model is drawn to generate the same sentence compared to the original sentence, which does not create a syntactically diverse paraphrase. To cope with this problem, we further propose using Virtual Adversarial Training (VAT) \citep{NEURIPS2019_812b4ba2} and noise perturbation added directly to the input embedding to steer the model towards a more diverse paraphrase generation, as in Figure \ref{fig:detail-lampat-architecture}. In addition, Large Language Models (LLMs), are known to experience the catastrophic forgetting \citep{kaushik2021understanding} during full fine-tuning; therefore, we adapt LoRA \citep{hu2021lora} as a parameter-efficient fine-tuning method to partially update the model's prior knowledge and preserve all the linguistic knowledge on which the model has been pre-trained. In general, LAMPAT can effectively generate human-like paraphrases while preserving the original semantic meaning and employing different syntactic structures to promote the diversity of the predictions.

In summary, the key contributions of this paper are as followed:
\begin{itemize}
    \item To resolve the requirement of parallel corpora for machine translation, we propose the unsupervised learning method for multilingual paraphrasing.
    \item To address the issue of predominantly generating identical or highly lexical-similar outputs, we incorporate noise perturbation and a virtual labeling strategy into the adversarial training process, aiming to alleviate this limitation.
    \item We expand the multilingual paraphrasing evaluation dataset to include more languages and leverage future research in multilingual paraphrase generation.
\end{itemize}

\section{Methodology}
\label{section:lampat}

The training process of our paraphrasing model is illustrated in Figure \ref{fig:lampat-architecture}, comprising three essential elements: Synthetic Parallel Corpora, Parameter-Efficient Fine-Tuning (PEFT), and Virtual Adversarial Training (VAT). We employ the Self-supervised model to generate paraphrases, which not only addresses the data shortage issue using unsupervised learning but also maintains semantic similarity and enhances lexical diversity as well.

\subsection{Problem Definition}
Given 2 sentences $x$ and $y$, where $x$ is the original sentence, and $y$ is the paraphrase reference of $x$. Let $\mathcal{M}(x)$ be the meaning of $x$ and $\mathcal{S}(x,y)$ be the lexical or syntactic similarity between $x$ and $y$. $\hat{y} = \underset{y}{argmax}[p(y|M(x)) - S(x,y)]$. The term $p(y|M(x))$ is the probability of generating $y$ that conveys the same meaning as $x$ (i.e. $M(x)$). $S(x,y) \in [0,1]$ measures the lexical similarity of $x$ and $y$, where $S(z,z)= 1$ for every $z$. Based on this formulation, paraphrasing should be a method which not only allows us to convey the same meaning but also enhances lexical diversity in the generated text. 

\subsection{Synthetic Parallel Corpora}
Synthetic Parallel Corpora is a significant component of LAMPAT. First, we corrupt the input by removing the stop words in the sentence, and then randomly shuffling the words in the remaining text. The corrupted sentence is referred to as the source sequence $S$, while the original uncorrupted sentence is the target sequence $T$. We have a set of stop words $A$, which are removed from the sentences. Our goal is to generate the paraphrase by reconstructing $T$ from the keywords or the corrupted sentence $S$, where $S = \text{Shuffle} (T - A)$. When we fine-tune the model, we create the input sequence $X$ by combining the source and target sequences with a special symbol in between. The input sequence $X$ is represented as $(x_1, x_2, x_3, ..., x_k$, \textit{\textbackslash n}, $x_{k+1}, x_{k+2}, ...x_h)$, where the source sequence is denoted as $S = (x_0, x_1, ..., x_k)$, and the target sequence is denoted as $T = (x_{k+1}, x_{k+2}, ..., x_h)$. The special character \textit{\textbackslash n} is included to differentiate between the source and target tokens, and it also serves as a prompt during the inference process.

\subsection{Parameter-Efficient Fine-Tuning}
The method chosen for parameter-efficient fine-tuning is Low-rank Adaptation (LoRA) \citep{hu2021lora}. A major drawback of fine-tuning is that the resulting model contains the same number of parameters as the original model. LoRA overcomes this by indirectly training some dense layers in a neural network by optimizing rank decomposition matrices of the dense layers' changes during adaptation while keeping the pre-trained weights frozen. This also allows for efficient task-switching by simply replacing the matrices, resulting in reduced storage requirements and task-switching overhead.

LoRA \citep{hu2021lora} introduces a constraint on the update of a pre-trained matrix $W_x \in \mathbb{R}^{d \times k}$ using a low-rank matrix $W_{\beta} \in \mathbb{R}^{d \times k}$. Instead of directly updating $W_x$, LoRA modifies it as $W'_x = W_x + W_{\beta}$ and focuses on updating the parameters involved in the construction of $W_{\beta}$. The construction of $W_{\beta}$ involves the multiplication of two matrices $B \in \mathbb{R}^{d \times r}$ and $A \in \mathbb{R}^{r \times k}$, where $r \ll min(d, k)$, resulting in a low-rank matrix $W'_x = W_x + BA$. By using LoRA, we reduce the number of parameters to tune from $d \times k$ to $r \times (d + k)$. Specifically, we follow the application of LoRA to the query and value transformation matrices in the multi-head attention sublayers, as in \citet{hu2021lora}.

\subsection{Virtual Adversarial Training}
Consider a standard classification task with an underlying data distribution $D$ over examples $x \in \mathbb{R}^d$ and corresponding labels $y$. Assume that we are given a suitable loss function $\mathcal{L}_{rec}$, for example, the cross-entropy loss for a neural network. As usual, $\theta \in \mathbb{R}^p$ is the set of model parameters. Our goal is then to find model parameters $\theta$ that satisfy:
\begin{equation}
    \min_{\theta}E_{(x,y)\sim \mathcal{D}}[\mathcal{L}_{rec}(f(x,\theta), y)]
    \label{eq:emperical_risk}
\end{equation}

Once we have created the synthetic parallel data, we define a set of permissible perturbations $\mathcal{C} \subseteq \mathbb{R}^d$ for each data point $x$, which represents the manipulative ability of the adversary. Instead of directly using samples from the distribution $\mathcal{D}$ in the loss $\mathcal{L}$, we allow the adversary $\delta \in \mathcal{C}$ to first perturb the input embedding. Specifically, $x$ denotes the sub-word embedding in $f(x, \theta)$. We notice that by perturbing the embedding space $x + \delta$, rather than the input space, adversarial training may unintentionally favour on-manifold perturbations over regular perturbations, leading to improved generalization. Hence, we apply perturbations to the embedding space.  On the other hand, complete label information may not always be available, and especially in this unsupervised manner, we aim to output virtual labels other than the label $y$. Consequently, we adopt a strategy to replace the label $y$ with its current approximation, $f(x, \theta)$. This approximation is not necessarily naive, as $f(x, \theta)$ tends to be close to $y$ when the number of labelled training samples is large. This rationale also explains the use of the term ``virtual'' in \citet{miyato2018virtual}. Essentially, we employ virtual labels generated from $f(x, \theta)$ in place of paraphrasing labels, and compute the adversarial direction based on these virtual labels. As a result, we replace the Equation \ref{eq:emperical_risk} by:
\begin{equation}
    \begin{split}
        \min_{\theta}E_{(x,y)\sim \mathcal{D}} \left[ \max_{\delta} \left[ \mathcal{L}_{rec}(f(x + \delta,\theta), y) \right. \right. \\
        \left. \left. + \alpha \mathcal{L}_{vadv}(f(x+\delta,\theta),f(x,\theta))  \right] \vphantom{\max_{\delta}} \right]
    \end{split}
    \label{eq:vat_adv_emperical_risk}
\end{equation}

\begin{algorithm*}[ht]
\fontsize{10pt}{10pt}\selectfont
\caption{Low-rank Adaptation Multilingual Paraphrasing using Adversarial Training.}
\label{alg:main_algo}
    \textbf{Input}: $X$: Training samples, $f(x; \theta)$: the machine learning model parameterized by $\theta$, $\epsilon$: the perturbation bound, $\tau$: the global learning rate, $\alpha$: the smoothing proportion of the adversarial training, $\eta$: the  ascent step size, $\mathcal{H}$: the Hessian gradient matrix, $N$: the number of epochs, $K$: the number of ascent steps, $e^*$: the number of epochs trained with PGD algorithm, $\gamma$: the scaling factor. \\
    \textbf{Output}: $\theta$.
    \begin{algorithmic}[1]
    \FOR{epoch = 1...$N$}
        \FOR{minibatch $B$ $\in$ X}
            \STATE $\delta \sim \gamma \cdot \mathcal{N}(0,\sigma^{2}I)$
            \\
            \FOR{m = 1...K}
                \STATE Accumulate gradient of parameter $\theta$:
                \\
                \STATE $g_{m} \leftarrow  g_{m-1} + \frac{1}{K} E_{(x,y) \in B} [\nabla_{\theta} \mathcal{L}_{rec}(f(x + \delta,\theta), y) +\alpha \nabla_{\theta}\mathcal{L}_{vadv}(f(x+\delta,\theta),f(x,\theta))]$
                \\
                \STATE Calculate the gradient of the perturbation $\delta$:
                \\
                \STATE $g_{adv} \leftarrow \nabla_{\delta}\mathcal{L}_{vadv}(f(x + \delta, \theta), f(x, \theta))$
                \\
                \STATE Update the perturbation $\delta$ through gradient ascent:
                \\
                \IF{epoch $\leq$ $e^*$}
                \STATE $\delta \leftarrow \prod_{||\delta|| \leq \epsilon} (\delta + \eta \frac{g_{adv}}{||g_{adv}||_F})$
                \ELSE
                \STATE $\delta \leftarrow \prod_{\mathcal{H}} (\delta + \eta [\mathcal{H}]^{-1} \frac{g_{adv}}{||g_{adv}||_F})$
                \ENDIF
            \ENDFOR
            \STATE Update the parameter $\theta$ through gradient descent:
            \STATE $\theta \leftarrow \theta - \tau g_{K}$
        \ENDFOR
    \ENDFOR
    \end{algorithmic}
\end{algorithm*}

In reference to Algorithm \ref{alg:main_algo}, our method can reduce the computational cost of adversarial training using projection over constraints algorithms. It achieves similar levels while conducting fewer sweeps of forward and backward propagation, making it faster and less computationally expensive. Our method takes advantage of every propagation to update weights and allows multiple updates per iteration, potentially leading to faster convergence. According to \citet{NEURIPS2019_812b4ba2}, by combining these factors, the $K$ ascent steps significantly accelerate standard adversarial training.  After that, the model's parameter $\theta$ is updated all at once with the accumulated gradients. By taking a descent step along the $K$ gradients, we can approximately calculate the following objective function:
\begin{equation}
    \begin{split}
        \min_{\theta}E_{x,y \sim \mathcal{D}} \left[ \frac{1}{K} \sum_{t=0}^{K-1} \max_{\delta} [ \mathcal{L}_{rec}(f(x + \delta,\theta), y) \right. \\
        \left. + \alpha \mathcal{L}_{vadv}(f(x+\delta,\theta),f(x,\theta)) ] \vphantom{\frac{1}{K}} \right]
    \end{split}
    \label{eq:main_emperical_risk}
\end{equation}

The rationale behind computing $g(\delta)$ with respect to the virtual adversarial regularizer $\nabla_{\delta} \mathcal{L}_{vadv}$ instead of $\nabla_{\delta} \mathcal{L}_{rec}$ in Algorithm \ref{alg:main_algo} is due to the unsupervised nature of the model training and the objective to guide the model towards the virtual labels rather than reconstructing the original sentence, which could result in duplication. Equation \ref{eq:main_emperical_risk} is essentially replacing the original batch $x$ with a virtual batch that is $K$ times larger, comprising samples with embeddings of $X + \delta_0,..., X + \delta_{K-1}$. While the original adversarial training Equation \ref{eq:vat_adv_emperical_risk} minimizes the maximum risk at a single estimated point in the vicinity of each training sample, Equation \ref{eq:main_emperical_risk} minimizes the maximum risk at each ascent step and guides the model towards the virtual labels with minimal additional overhead. Moreover, we use both Projected Gradient Descent (PGD) and Projected-Newton Method (PNM) in Algorithm \ref{alg:main_algo}. The Gradient Descent step involves descending along the linear estimate of the function, while Newton's step involves moving the point towards the minimum of the parabola that approximates the function, which can lead to faster convergence.

\section{Experimental Setup}
\subsection{Primary Model}

\textbf{Pre-trained model}: We utilize the mGPT model, a pre-trained multilingual GPT-like model, with a 1.3B checkpoint consisting of 100K tokens and supporting 61 languages. The LoRA implementation from PEFT \footnote{\url{https://github.com/huggingface/peft}}. 

\textbf{Dataset}: To assess the fine-tuning, we choose to use the latest version WMT19 \citep{wmt19translate} to train the model. This dataset covers a wide range of 15 languages including Arabic, Czech, German, English, Spanish, French, Hindi, Indonesian, Italian, Japanese, Kazakh, Dutch, Portuguese, Russian, and Chinese. The WMT19 dataset we use is in its latest version, which is just released in 2023. To balance language resources, we employ a uniform distribution to sample sentences, creating a training set of nearly 600k sentences and a validation set of around 100k sentences. For training, we use the monolingual version of WMT19 and corrupt the input sentence by removing all of the stop words \footnote{\url{https://github.com/stopwords-iso/stopwords-iso}}, further we randomly shuffle the words 33\% of the time. The goal is to reconstruct a sentence from its keywords, or its corrupted sentence. For the machine translation approach, we use the available bilingual version available in WMT19, we sample with the same strategy as LAMPAT's training with 600k samples for training and 100k for validation.

\subsection{Baseline Model}
We compare our method with other approaches such as multilingual machine translation (MMT) proposed in \citet{thompson-post-2020-paraphrase} and denoising auto-encoder (DAE) in \citet{guo2019zeroshot}. Initially, these methods are trained on different datasets, thus, we re-train them, following the procedures proposed in each paper, on the WMT19 dataset. 

\subsection{Evaluation Dataset}
We follow \citet{guo2019zeroshot}, randomly select 10k sentences respectively from each language of English, Spanish, Russian, and Chinese to construct the test set. However, the number of languages covered in \citet{guo2019zeroshot} is relatively small compared to the number of languages around the world, over 7000 languages \footnote{\url{https://www.ethnologue.com/statistics}}. Therefore, we expand the languages covered in the task of multilingual paraphrasing to 13 languages, including some figurative languages such as Japanese, and Chinese or accented languages such as Vietnamese. The proposed evaluation dataset has two types to assess different aspects of the model: \textbf{Input-only} and \textbf{Input-reference}.

\textbf{Input-only}: We follow \citet{thompson-post-2020-paraphrase} and use the validation set from the WMT19 dataset, which is realeased in 2019 and available on HuggingFace \footnote{\url{https://huggingface.co/datasets/wmt19}}. Since the WMT19 dataset is used for the task Machine Translation, we thus, extract one side of the dataset in order to build the \textbf{Input-only} evaluation dataset. The languages we extracted from the WMT19 are Czech (cs), German (de), English (en), Finish (fi), French (fr) and Chinese (zh). Most of the samples in this evaluation dataset are of the news domains, which is used to test the model's ability on producing a paraphrase that conveys the same meaning.

\textbf{Input-reference}: We use in total 3 datasets to construct this evaluation set:
\begin{itemize}
    \item \textbf{PAWS-X} \citep{yang-etal-2019-paws} is the cross-lingual paraphrase identification dataset, thus, we extract only the sentence pairs with label 1 (indicating paraphrase) for 6 languages: Japanese (ja), Chinese (zh), German (de), French (fr), Spanish (es) and English (en).
    
    \item \textbf{Opusparcus} \citep{Creutz2019}: is a paraphrase corpus for six European languages: German (de), English (en), Finnish (fi), French (fr), Russian (ru), and Swedish (sv).  We extract the test set of Opusparcus with a score of 4 to ensure high-quality sentence pairs.
    
    \item \textbf{STAPLE} \citep{DVN/38OJR6_2020}: is a multi-reference machine translation dataset in which each reference could be viewed as the paraphrase. Since STAPLE does not have the validation or test set, we randomly extract 1000 samples, each with 5 reference texts, covering 3 languages: Vietnamese (vi), Portuguese (pt) and Hungarian (hu), to construct the first multi-reference paraphrase generation corpus with 1000 samples and 4 references each.
\end{itemize}

\subsection{Automatic Evaluation}
We follow \citet{Chowdhury_Zhuang_Wang_2022} to report the result on BLEU \citep{papineni-etal-2002-bleu}, Self-BLEU, Self-TER, which adapted TER \citep{snover-etal-2006-study} to the input instead of the reference, BERTScore \citep{zhang-etal-2020-bertscore} with \texttt{bert-base-multilingual-cased} checkpoint, iBLEU with $\alpha = 0.7$ following \citet{hosking-lapata-2021-factorising}. In addition, we further use two latest paraphrase metrics: ParaScore \citep{shen-etal-2022-evaluation} and BERT-iBLEU \citep{niu-etal-2021-unsupervised}.

\subsection{Human Evaluation}
In addition to machine evaluation, we also conduct the human evaluation of the paraphrase generated by our model. For each of the following languages: English, Vietnamese, German, French and Japanese, we randomly extract 200 sentence triples of the input sentence, our model prediction and the output from the model of \citet{thompson-post-2020-paraphrase}. For each mentioned language, we ask 5 annotators to score 200 sentence pairs independently. Each annotator is instructed to rate on the 1-5 scale (with 5 being the highest) based on 3 criteria: \textbf{(i) Semantic preservation}, evaluating how much information is preserved in the output; \textbf{(ii) Lexical similarity}, evaluating how much similar in term of syntax or word choices of the output compared to input; and \textbf{(iii) Fluency}, assessing the fluency and coherence of the generated output. The annotators' agreement is measured using Krippendorff's alpha \citep{Krippendorff1970EstimatingTR}, which provides a measure of inter-annotator reliability.

\section{Results}
\subsection{Main Results}

\begin{table}[h!]
    \centering
    \fontsize{9pt}{9pt}\selectfont
    \begin{tabular}{|l|rrrr|} 
    \hline
\multicolumn{1}{|c|}{Method} & \multicolumn{1}{c}{en} & \multicolumn{1}{c}{es} & \multicolumn{1}{c}{zh} & \multicolumn{1}{c|}{ru} \\ \hline
\multicolumn{5}{|c|}{\textit{BERTScore} $\uparrow$}                                                                                                                                 \\ \hline
DAE                           & 79.25                           & 80.56                           & 78.91                           & 77.83                           \\
MMT                         & 84.93                           & 82.68                           & \textbf{84.79}                  & 81.32                           \\
LAMPAT                   & \textbf{86.86}                  & \textbf{84.35}                  & 83.26                           & \textbf{84.01}                  \\ \hline
\multicolumn{5}{|c|}{\textit{Self-BLEU} $\downarrow$}                                                                                                                                 \\ \hline
DAE                           & 20.35                           & 30.49                           & 20.38                           & 20.61                           \\
MMT                         & 19.89                           & 28.57                           & \textbf{10.32}                  & 15.19                           \\
LAMPAT                   & \textbf{19.46}                  & \textbf{20.16}                  & 14.95                           & \textbf{12.46}                  \\ \hline
\multicolumn{5}{|c|}{\textit{Self-TER} $\uparrow$}                                                                                                                                  \\ \hline
DAE                           & 50.48                           & 45.19                           & 45.92                           & 48.31                           \\
MMT                         & 52.45                           & 43.16                           & 41.25                           & 45.68                           \\
LAMPAT                   & \textbf{61.32}                  & \textbf{55.43}                  & \textbf{55.28}                  & \textbf{50.67}                  \\ \hline
\multicolumn{5}{|c|}{\textit{BERT-iBLEU} $\uparrow$}                                                                                                                                                     \\ \hline
DAE                           & 79.33                           & 78.08                           & 79.05                           & 78.14                            \\
MMT                         & 83.92                           & 80.16                           & \textbf{85.72}                           & 81.99                            \\
LAMPAT                   & \textbf{85.52}                  & \textbf{83.41}                  & 83.61                  & \textbf{84.69}                   \\ \hline
\multicolumn{5}{|c|}{\textit{ParaScore} $\uparrow$}                                                                                                                                 \\ \hline
DAE                           & 88.75                           & 90.46                           & 89.37                           & 88.50                           \\
MMT                         & 89.45                           & 91.56                           & 90.05                           & 88.47                           \\
LAMPAT                   & \textbf{92.95}                  & \textbf{92.96}                  & \textbf{90.64}                  & \textbf{91.92}    \\\hline             
\end{tabular}
\caption{Multilingual paraphrase generation test results over 4 languages English, Spanish, Chinese and Russian from the work of DAE \citep{guo2019zeroshot}.}
\label{tab:main_result}
\end{table}

\begin{table}[h!]
    \centering
    {\fontsize{9pt}{9pt}\selectfont
    \begin{tabular}{|l|cccccc|}
\hline
\multicolumn{1}{|c|}{Method} &
  \multicolumn{1}{c}{fr} &
  \multicolumn{1}{c}{cs} &
  \multicolumn{1}{c}{fi} &
  \multicolumn{1}{c}{de} &
  \multicolumn{1}{c}{en} &
  \multicolumn{1}{c|}{zh} \\ \hline
\multicolumn{7}{|c|}{BERTScore $\uparrow$}                                                                                                                \\ \hline
DAE               & 70.39          & 78.45          & 80.96          & 69.32          & 65.68          & \textbf{81.35} \\
MMT & 74.40          & 80.20          & 81.20          & 71.20          & 65.48          & 80.29          \\
LAMPAT                 & \textbf{85.42} & \textbf{83.92} & \textbf{84.91} & \textbf{86.16} & \textbf{82.59} & 79.18          \\ \hline
\multicolumn{7}{|c|}{Self-TER $\uparrow$}                                                                                                                 \\ \hline
DAE               & 45.53          & 39.97          & 54.50          & 46.18          & 30.42          & 28.49          \\
MMT & 48.40          & 37.12          & 56.74          & 48.42          & 33.76          & 31.56          \\
LAMPAT                 & \textbf{49.53} & \textbf{40.49} & \textbf{63.92} & \textbf{53.76} & \textbf{39.31} & \textbf{40.80} \\ \hline
\multicolumn{7}{|c|}{BERT-iBLEU $\uparrow$}                                                                                                               \\ \hline
DAE               & 71.10          & 79.56          & 81.38          & 71.24          & 67.36          & \textbf{81.05} \\
MMT & 75.48          & 78.81          & 82.15          & 74.16          & 66.08          & 69.70          \\
LAMPAT                 & \textbf{84.63} & \textbf{85.22} & \textbf{86.03} & \textbf{85.38} & \textbf{84.08} & 79.90          \\ \hline
\multicolumn{7}{|c|}{ParaScore $\uparrow$}                                                                                                                \\ \hline
DAE               & 89.91          & 88.46          & 75.93          & 87.43          & 88.56          & 88.12          \\
MMT & 89.95          & 90.10          & 81.15          & 89.47          & 91.49          & 90.35          \\
LAMPAT                 & \textbf{93.24} & \textbf{92.45} & \textbf{86.82} & \textbf{93.21} & \textbf{94.84} & \textbf{92.24} \\ \hline
\end{tabular}}
\caption{Multilingual paraphrase generation test results on our input-only evaluation dataset.}
\label{tab:perf_inp_only}
\end{table}

\begin{table*}[h!]
    \centering
    {\begin{tabular}{|lrrrrrrrrrrrr|}
\hline
\multicolumn{1}{|c|}{Method} &
  \multicolumn{1}{c}{sv} &
  \multicolumn{1}{c}{fi} &
  \multicolumn{1}{c}{en} &
  \multicolumn{1}{c}{fr} &
  \multicolumn{1}{c}{de} &
  \multicolumn{1}{c}{ru} &
  \multicolumn{1}{c}{ja} &
  \multicolumn{1}{c}{zh} &
  \multicolumn{1}{c}{es} &
  \multicolumn{1}{c}{hu} &
  \multicolumn{1}{c}{pt} &
  \multicolumn{1}{c|}{vi} \\ \hline
\multicolumn{13}{|c|}{BERTScore $\uparrow$} \\ \hline
\multicolumn{1}{|l|}{DAE} &
  80.20 &
  79.50 &
  84.00 &
  85.20 &
  89.44 &
  89.90 &
  74.50 &
  \textbf{83.00} &
  88.80 &
  75.30 &
  78.40 &
  80.50 \\
\multicolumn{1}{|l|}{MMT} &
  83.48 &
  84.92 &
  75.86 &
  77.01 &
  82.53 &
  81.49 &
  72.61 &
  71.93 &
  82.45 &
  73.11 &
  83.44 &
  75.76 \\
\multicolumn{1}{|l|}{LAMPAT} &
  \textbf{85.47} &
  \textbf{85.47} &
  \textbf{94.87} &
  \textbf{89.94} &
  \textbf{90.10}&
  \textbf{92.16} &
  \textbf{90.77} &
  81.00 &
  \textbf{92.87} &
  \textbf{86.16} &
  \textbf{91.99} &
  \textbf{87.70} \\ \hline
\multicolumn{13}{|c|}{BLEU $\uparrow$} \\ \hline
\multicolumn{1}{|l|}{DAE} &
  5.22 &
  4.20 &
  10.50 &
  15.51 &
  14.44 &
  5.46 &
  20.62 &
  \textbf{47.80} &
  18.58 &
  8.10 &
  16.01 &
  15.05 \\
\multicolumn{1}{|l|}{MMT} &
  1.22 &
  0.39 &
  11.79 &
  6.10 &
  11.98 &
  0.77 &
  9.55 &
  16.69 &
  20.49 &
  0.84 &
  11.09 &
  3.58 \\
\multicolumn{1}{|l|}{LAMPAT} &
  \textbf{6.04} &
  \textbf{4.90} &
  \textbf{23.07} &
  \textbf{20.55} &
  \textbf{21.65} &
  \textbf{6.48} &
  \textbf{29.13} &
  30.52 &
  \textbf{26.95} &
  \textbf{8.16} &
  \textbf{16.65} &
  \textbf{20.24} \\ \hline
\multicolumn{13}{|c|}{Self-BLEU $\downarrow$} \\ \hline
\multicolumn{1}{|l|}{DAE} &
  24.58 &
  17.56 &
  45.50 &
  30.20 &
  25.75 &
  36.82 &
  50.78 &
  50.22 &
  40.69 &
  25.82 &
  \textbf{30.44} &
  27.26 \\
\multicolumn{1}{|l|}{MMT} &
  25.72 &
  16.76 &
  50.33 &
  23.47 &
  35.65 &
  39.67 &
  45.62 &
  63.32 &
  36.65 &
  27.82 &
  34.70 &
  27.24 \\
\multicolumn{1}{|l|}{LAMPAT} &
  \textbf{23.68} &
  \textbf{14.53} &
  \textbf{43.47} &
  \textbf{19.55} &
  \textbf{24.54} &
  \textbf{30.25} &
  \textbf{40.56} &
  \textbf{47.58} &
  \textbf{29.00} &
  \textbf{22.94} &
  31.43 &
  \textbf{25.51} \\ \hline
\multicolumn{13}{|c|}{iBLEU $\uparrow$} \\ \hline
\multicolumn{1}{|l|}{DAE} &
  0.05 &
  0.01 &
  -0.04 &
  -0.05 &
  -0.01 &
  0.01 &
  0.08 &
  0.10 &
  0.04 &
  0.05 &
  0.01 &
  -0.01 \\
\multicolumn{1}{|l|}{MMT} &
  0.01 &
  0.02 &
  0.02 &
  \textbf{0.04} &
  0.04 &
  -0.03 &
  0.06 &
  0.12 &
  0.09 &
  0.06 &
  0.11 &
  \textbf{0.14} \\
\multicolumn{1}{|l|}{LAMPAT} &
  \textbf{0.08} &
  \textbf{0.04} &
  \textbf{0.04} &
  0.02 &
  \textbf{0.05} &
  \textbf{0.04} &
  \textbf{0.09} &
  \textbf{0.15} &
  \textbf{0.13} &
  \textbf{0.15} &
  \textbf{0.15} &
  0.05 \\ \hline
\multicolumn{13}{|c|}{Self-TER $\uparrow$} \\ \hline
\multicolumn{1}{|l|}{DAE} &
  50.18 &
  \textbf{60.57} &
  24.55 &
  55.34 &
  57.44 &
  33.14 &
  40.27 &
  30.75 &
  31.55 &
  40.57 &
  47.81 &
  50.55 \\
\multicolumn{1}{|l|}{MMT} &
  47.16 &
  53.59 &
  23.90 &
  47.07 &
  43.48 &
  29.27 &
  40.88 &
  26.85 &
  30.97 &
  36.88 &
  27.10 &
  39.95 \\
\multicolumn{1}{|l|}{LAMPAT} &
  \textbf{57.20} &
  59.45 &
  \textbf{56.56}&
  \textbf{56.49} &
  \textbf{61.33} &
  \textbf{47.80} &
  \textbf{42.62}&
  \textbf{35.34} &
  \textbf{38.55} &
  \textbf{41.51} &
  \textbf{55.11} &
  \textbf{57.91} \\ \hline
\multicolumn{13}{|c|}{BERT-iBLEU $\uparrow$} \\ \hline
\multicolumn{1}{|l|}{DAE} &
  70.20 &
  78.20 &
  66.60 &
  56.20 &
  62.70 &
  68.20 &
  52.00 &
  67.20 &
  57.20 &
  67.50 &
  68.80 &
  77.20 \\
\multicolumn{1}{|l|}{MMT} &
  61.43 &
  68.44 &
  65.25 &
  66.33 &
  63.07 &
  69.06 &
  65.33 &
  62.67 &
  63.95 &
  72.39 &
  72.34 &
  69.91 \\ 
\multicolumn{1}{|l|}{LAMPAT} &
  \textbf{73.30} &
  \textbf{78.60} &
  \textbf{67.60} &
  \textbf{82.33} &
  \textbf{80.25} &
  \textbf{69.98} &
  \textbf{78.04} &
  \textbf{67.72} &
  \textbf{86.46} &
  \textbf{76.32} &
  \textbf{78.92} &
  \textbf{81.39} \\ \hline
\multicolumn{13}{|c|}{ParaScore $\uparrow$} \\ \hline
\multicolumn{1}{|l|}{DAE} &
  82.00 &
  72.80 &
  85.00 &
  85.30 &
  88.90 &
  90.20 &
  84.50 &
  87.60 &
  90.10 &
  80.50 &
  80.52 &
  80.02 \\
\multicolumn{1}{|l|}{MMT} &
  83.17 &
  81.72 &
  76.83 &
  77.62 &
  82.59 &
  81.69 &
  73.92 &
  73.07 &
  82.70 &
  74.54 &
  83.84 &
  77.01 \\
\multicolumn{1}{|l|}{LAMPAT} &
  \textbf{86.42} &
  \textbf{86.00} &
  \textbf{94.51}&
  \textbf{90.73} &
  \textbf{90.29} &
  \textbf{91.47} &
  \textbf{91.33} &
  \textbf{93.01} &
  \textbf{92.96} &
  \textbf{86.89} &
  \textbf{91.74} &
  \textbf{89.78} \\ \hline
\end{tabular}}
\caption{Multilingual paraphrase generation test results on our input-reference evaluation dataset.}
\label{tab:perf_inp_ref}
\end{table*}

For the test dataset from \citet{guo2019zeroshot} and our \textbf{input-only} evaluation, we evaluate using BERTScore, Self-BLEU, Self-TER, BERT-iBLEU and ParaScore, as these evaluation metrics do not require the reference text. The test results are depicted by Table \ref{tab:main_result} and \ref{tab:perf_inp_only}. For \textbf{input-reference}, we report BLEU and iBLEU, in addition, as depicted by Table \ref{tab:perf_inp_ref}. LAMPAT can generate diverse output compared to the input, which is indicated by the low score in Self-BLEU and high score in Self-TER. LAMPAT also preserves better information from the input as demonstrated by the BERTScore results. We have manually examined the text generated by all four methods by randomly selecting 100 samples per language. Although these sentences convey the same meaning, the word choices and syntax structures are largely different from both input and reference, leading to the low score of iBLEU. Overall, BERT-iBLEU and ParaScore metrics, which are the metrics that grade both lexical and semantic aspects, show that all three methods could generate comprehensive sentences. However, our method, LAMPAT, still achieves the highest score over 13 languages we have tested. Even though our model learned in an unsupervised manner, it can still outperform the supervised counterpart. Overall, LAMPAT has demonstrated that integrating adversarial training into unsupervised learning could improve the performance of multilingual paraphrase generation.

\subsection{Human Evaluation Results}
The results of our human evaluation can be found in Table \ref{tab:human_eval}, where we present the assessments provided by human evaluators. The evaluations conducted by human experts provide valuable insights into the performance of our model.

\begin{table}[ht]
    \centering
    {\fontsize{9pt}{9pt}\selectfont\begin{tabular}{|l|rrr|}
        \hline
        \multicolumn{1}{|c|}{Method} & SP $\uparrow$ & LS $\downarrow$ & F $\uparrow$\\ \hline
        MMT & 3.2 & 4.8 & 4.8 \\
        LAMPAT & 4.2 & 3.2 & 4.8 \\ \hline
        Human generated paraphrase & 4.4 & 2.4 & 4.8 \\ \hline \hline
        Krippendorff's alpha & 0.68 & 0.7 & 0.78 \\ \hline
    \end{tabular}}
    \caption{Human evaluation results. SP: Semantic Preservation; LS: Lexical Similarity; F: Fluency. All the scores reported are the average value of 5 chosen languages.}
    \label{tab:human_eval}
\end{table}

\section{Ablation Study}
\label{sec:ablation_study}
\subsection{Parameter-Efficient Fine-Tuning}
In order to examine which ingredients help improve the performance of LAMPAT, we experiment with Adapter \citep{pmlr-v97-houlsby19a}, Prefix Tuning \citep{li-liang-2021-prefix}, Prompt Tuning \citep{lester-etal-2021-power}, LoRA \citep{hu2021lora}, P-Tuning \citep{liu-etal-2022-p} and Full fine-tuning to find out which PEFT methods result in a more stable and better result. For each of the methods, we train on the same number of epochs with 3 random seeds and report the mean and standard deviation of ParaScore of all languages.
\begin{table}[h!]
\centering
    {\fontsize{9pt}{9pt}\selectfont\begin{tabular}{|l|c|}
        \hline
        \multicolumn{1}{|c|}{Method} & ParaScore \\ \hline
        Prefix Tuning & 77.45$\pm$8.5 \\
        Prompt Tuning & 82.25$\pm$7.9 \\
        P-Tuning & 82.94$\pm$5.5 \\
        Adapter & 89.56$\pm$3.2 \\
        LoRA & \textbf{90.67$\pm$2.5} \\ 
        Full & 89.46$\pm$1.5 \\ \hline
    \end{tabular}}
\caption{ParaScore of different fine-tuning methods on 13 languages. The mean and standard deviation are the weighted mean and standard deviation in all 13 languages.}
\label{tab:ablation_study}
\end{table}

According to Table \ref{tab:ablation_study}, LoRA experiences to be the most stable method and achieves the highest scores, especially for generation tasks. 

\subsection{Adversarial Optimization}
\begin{figure}[h!]
    \centering
    \includegraphics[width=0.8\linewidth]{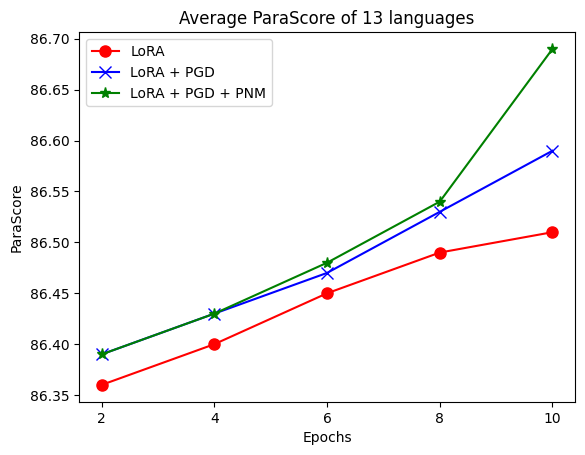}
    \caption{The average ParaScore of each technique over 13 languages.}
    \label{fig:avg_records_of_adv}
\end{figure}

Since LoRA is a stable method for partially fine-tuning LLMs, we adapt LoRA as the PEFT method for fine-tuning mGPT. In order to study the effect of adversarial training in unsupervised learning, we employ two settings which are Projected Gradient Descent (PGD) and Projected Newton Method (PNM) \cite{Bertsekas1982PNM} together with LoRA. According to Figure \ref{fig:avg_records_of_adv}, we hypothesized that the main driving force that makes LoRA + PGD + PNM better compared to PGD is because it uses the Taylor expansion, which has a better approximation of the objective function. In general, when we are near the local optima (for example, at the last 2 epochs), we can take a few more Newton's steps to reach the optimum point instead of taking many small Gradient Descent steps. 

\section{Related Works}

\subsection{Multilingual Paraphrasing}
Numerous techniques for paraphrasing in multiple languages employ Machine Translation-based models (MNMT). To illustrate, \citet{thompson-post-2020-paraphrase} applied a pretrained MNMT model introduced by \citet{thompson-post-2020-automatic}, along with a customized decoding algorithm aimed at reducing repetition of words and encouraging diverse vocabulary usage. Another instance is the work by \citet{guo2019zeroshot}, which leveraged a language model pretraining task adapted from \citet{NEURIPS2019_c04c19c2}. During inference, the same language code is provided to the model, and the sequence is generated sequentially in an autoregressive manner. Despite the fact that the translation-based approach produces high-quality and fluent paraphrases, it faces certain inherent challenges. Firstly, there is a potential for bias stemming from the dominance of certain languages used for training the MNMT model. Secondly, there may be instances of incorrect synthetic paraphrases due to the inherent ambiguity in the pivot language, as pointed out by \citet{thompson-post-2020-paraphrase}. These issues need careful consideration in the development of multilingual paraphrasing methods.

\subsection{Adversarial Training}
Adversarial training is a powerful technique employed in the development of resilient neural networks. While the computer vision community, as highlighted by \citet{goodfellow2015explaining}, has generally accepted that adversarial training can have a detrimental impact on model generalization, the scenario appears to be quite different for language models, as evidenced by studies such as \citet{pereira-etal-2020-adversarial} and \citet{dong2021towards}. Incorporating adversarial training into large language models (LLMs), as explored by \citet{miyato2018virtual} and further elaborated upon by \citet{dong2021should}, has been found to yield improvements in both model generalization and robustness. An innovative training algorithm, denoted as YOPO (\textbf{Y}ou \textbf{O}nly \textbf{P}ropagate \textbf{O}nce), was proposed by \citet{NEURIPS2019_812b4ba2}. YOPO takes advantage of the ``free'' training strategies advocated by \citet{NEURIPS2019_7503cfac} to diversify the training data by incorporating various adversarial samples while imposing different norm constraints. Also, \citet{miyato2018virtual} proposes a new training method that regularizes the training objective by using virtual labels in adversarial training. These approaches collectively showcase the effectiveness of adversarial training in enhancing both the robustness and generalization capabilities of neural networks.

\section{Conclusion and Future Work}
In this research, we introduce an efficient method for generating paraphrases in multiple languages using Low-Rank Adaptation combined with virtual labeling during adversarial training. Importantly, our approach delivers satisfactory results without relying on supervised learning. Additionally, we contribute to the field by creating a novel multilingual multi-domain evaluation dataset. While LAMPAT has demonstrated proficiency in generating human-like paraphrases across various languages, it still requires improvements in handling idiomatic expressions. Furthermore, our evaluation dataset covers only 13 languages, leaving out many, especially those with limited resources. This highlights the ongoing need for research to enhance and expand the capabilities of multilingual paraphrase generation models.

\section{Acknowledgement}
This research/project is supported by the National Research Foundation, Singapore under its AI Singapore Programme, AISG Award No: AISG2-TC-2022-005.

\bibliography{anthology,custom}

\end{document}